\documentclass{article}
\usepackage{multirow}

\usepackage[a4paper,margin=1in]{geometry}
\usepackage{setspace}
\usepackage{graphicx}
\usepackage{natbib}
\usepackage{hyperref}
\usepackage{lineno}
\usepackage{tabularx}
\usepackage{array}

\newcommand{\texthash}{\#}

\title{If They Disagree, Will You Conform? \\
Exploring the Role of Robots' Value Awareness in a Decision-Making Task
}

\author{
    Giulia Pusceddu$^{1\ast}$,
    Giulio Antonio Abbo$^{2}$,\\
    Francesco Rea$^{1}$,
    Tony Belpaeme$^{2}$, 
    Alessandra Sciutti$^{1}$
    \and
	\small$^{1}$CONTACT Unit, Italian Institute of Technology, Genoa, Italy.\and
	\small$^{2}$IDLab-AIRO, Ghent University -- imec, Belgium\and
	\small$^\ast$Corresponding author. Email: giulia.pusceddu@iit.it\and
}
\date{\textit{This is the pre-print version of the manuscript, before peer-review. The peer-reviewed version of this manuscript was accepted for publication in \textbf{Interaction Studies} (John Benjamins Publishing Company).}
\textit{The final version will be available at:} \\
\url{https://benjamins.com/catalog/is}}

\begin{document}
\maketitle

\begin{abstract}
\noindent
This study investigates whether the opinions of robotic agents can influence human decision-making when robots display value awareness (i.e., the capability of understanding human preferences and prioritizing them in decision-making). 
We designed an experiment in which participants interacted with two Furhat robots – one programmed to be Value-Aware and the other Non-Value-Aware – during a labeling task for images representing human values. 
Results indicate that participants distinguished the Value-Aware robot from the Non-Value-Aware one. 
Although their explicit choices did not indicate a clear preference for one robot over the other, participants directed their gaze more toward the Value-Aware robot. 
Additionally, the Value-Aware robot was perceived as more loyal,
suggesting that value awareness in a social robot may enhance its perceived commitment to the group.
Finally, when both robots disagreed with the participant, conformity occurred in about one out of four trials, and participants took longer to confirm their responses, suggesting that two robots expressing dissent may introduce hesitation in decision-making. 
On one hand, this highlights the potential risk that robots, if misused, could manipulate users for unethical purposes. On the other hand, it reinforces the idea that social robots could encourage reflection in ambiguous situations and help users avoid scams.
\end{abstract}

\noindent\textbf{Keywords:} Value Awareness, Conformity, Robot Influence, Decision-Making, Multi-party Interaction, Group Human-Robot Interaction, Group Dynamics 

\section{Introduction \& Motivation}
Advancements in robotics and natural language processing and generation are leading to the deployment of robotic agents in social contexts, including education \citep{belpaeme2018social, belpaeme2021social, cocchella2023school} and elderly or hospital care \citep{cifuentes2020social, andtfolk2022humanoid}, where they could serve as tutors or assistants.
As robotic agents in these roles may need to support humans in complex decision-making processes and act as trustworthy social companions, evaluating trust toward them becomes crucial.

In human interactions, people tend to rely on those they perceive as competent and who share their motives \citep{twyman2008trust}. 
Similarly, trust in robots may depend not only on their perceived performance \citep{hancock2011meta} but also on their alignment with human values and goals, as recent research suggests \citep{bhat2024evaluating}. 
While studies have explored mechanisms underlying trust in social robots, little is known about how value alignment may influence human decision-making.

At the same time, establishing trust relationships with robotic agents may represent a safety risk, because they could be used to manipulate individuals through social engineering strategies \citep{aroyo2018trust, abate2020social}. 
This makes it crucial to anticipate and investigate these risks, particularly in situations where robots may exert social pressure or push to conformity. 

Building on previous research on robots' influence 
and conformity
in human-robot interactions, this study explores whether the opinions of robotic agents may affect human decision-making, introducing the novel aspect of robots' value awareness in the scenario.
Specifically,  we design the robots’ behavior to act as either Value-Aware or Non-Value-Aware and assess whether participants perceive this difference, during a task that involves labeling image sets representing human values.
We examine whether participants are more likely to conform to the suggestions of a value-aware robot compared to a non-value-aware one.
Additionally, we investigate whether a human participant conforms to two robotic agents. 
This study's objectives can thus be summarized as follows:
\begin{itemize}
    \item[MC] Manipulation check: Ensure that the participants distinguish the Value-Aware robot from the Non-Value-Aware one.
    \item[H1] Hypothesis 1: Participants are more likely to rely on the Value-Aware robot rather than the Non-Value-Aware one.
    \item[H2] Hypothesis 2: Participants conform to a group of two robots in a value-related task.
\end{itemize}


\noindent
To achieve these goals, the study is structured into two phases:
\begin{enumerate}
    \item \textbf{Stimuli Selection}: An online questionnaire to select image sets and keywords for use as stimuli in the experimental phase.
    \item \textbf{Experiments with Robotic Agents}: A series of experiments in which human participants interact with two Furhat robots to associate keywords with the image sets selected in the previous phase.
\end{enumerate}

\section{Related Work}
\subsection{Values in Human-Robot Interaction} 
Values are deeply held beliefs closely tied to emotions and serve as guiding principles in humans' everyday lives \citep{schwartz2012overview}.
There is no unified definition of values; instead, there are many frameworks, such as the theory of basic human values \citep{schwartz2012overview}, the moral foundations theory \citep{graham2013moral}, and the Rokeach value survey \citep{rokeach1973nature}.

If defining values is already complex for humans, in social robotics, the concept of value-aware robots is even harder to define. 
According to \cite{abbo2023users}, a value-aware robot should be capable of understanding human principles and prioritizing them in its decision-making.

Considering human-robot interactions, a study exploring users' perspectives on value awareness in social robots through a series of focus groups found that people are largely unaware of the potential misalignment between a robot’s behavior and their own values, and therefore remain unconcerned about it \citep{abbo2023users}. 
However, ensuring that robotic and virtual agents can align with human values is crucial for maintaining their behavior within moral bounds \citep{abbo2025concerns}.
Findings of a previous study \citep{ciupinskalabs}, in which participants interacted with the iCub robot in scenarios involving violations of privacy, freedom, and social norms, showed that participants judged privacy violations with moderate disagreement, whereas misbehaviors regarding social norms were least accepted.
These results, along with providing insights into which values may be most significant in interactions with a humanoid robot, underline the need for context-aware robotic behavior \citep{chevalier2022context}. 
An additional study suggests that people’s acceptance of a robot’s decisions depends on their perception of its ability to make value-aligned choices \citep{bhat2024evaluating}. 

Taking these considerations into account, value awareness appears to enhance human-robot interactions. 
However, shared guidelines for designing value-aware robots are still in the definition phase.
Building on the study by \cite{van2023candide}, one way to achieve value awareness in robots could be through narrative-based reasoning. 
They propose a computational model in which prior knowledge and beliefs help shape individual narratives.
As belief systems evolve through continuous experiences, narratives serve as a means to interpret and integrate new information.
Building on this idea, in our study, we designed a robot that demonstrates value awareness by associating human experiences with image descriptions. 
This approach has the goal of creating the impression that the robot's reasoning is grounded in a belief system, suggesting an underlying framework of values.

\subsection{Conformity and Influence in Human-Robot Teams}
In the last decade, several studies have attempted to clarify whether the influence of robots in decision-making scenarios mirrors that of humans.
Human-robot interaction studies have adapted variations of the Asch experiment \citep{asch1951group} to test conformity, defined as the act of changing one’s behavior to match the responses of others \citep{cialdini2004social}. 
However, findings have been inconsistent: while some studies report that human participants conform to a group of robots \citep{qin2022adults, salomons2018humans}, others do not observe this effect \citep{brandstetter2014peer, shiomi2016synchronized}.

Several factors may contribute to these discrepancies. 
Age appears to play a role, with younger participants exhibiting a higher tendency to conform \citep{vollmer2018children}. 
Task type influences conformity as well. For instance, previous research suggests that artificial agents are more trusted in analytical tasks than in social ones \citep{hertz2019mixing}. 
According to \cite{brandstetter2014peer}, task ambiguity might be another key factor: when participants evaluate comparable responses, they are more likely to conform to others’ opinions.

Other research suggests that in decision-making tasks, robot advice is primarily followed when perceived as low-impact \citep{sembroski2017he}, and that human team members tend to prefer following human peers over robots \citep{zhang2021you, pusceddu2025exploring}. 
Furthermore, studies examining the robot’s role within a group indicate that even when robots are introduced as leaders, they may not be perceived as such by human participants \citep{alves2016role}.

The role of robots in decision-making processes remains uncertain, and further research is needed to determine the conditions under which robots can effectively influence group dynamics. 
To address this gap, our study replicates an Asch-like task to assess human conformity within a team of two robots, focusing on an ambiguous task related to human values.

\section{Stimuli Selection}
\subsection{Participants}
Sixty-five volunteers (23 females, 37 males; $\mu_{age}=26.2$, $\sigma_{age}=5.9$ years) completed the online questionnaire generated using SoSci Survey \citep{leiner2024sosci}. 
They were recruited through a mailing list and by sharing the questionnaire link on social media.

The starting page of the survey informed that the collected data were anonymous, would be analyzed in aggregate form for scientific purposes only, and that proceeding to the form completion implied agreement to these terms.

The administration of this survey is compliant with the ethical protocol IIT\_INT\_HRI\_rev03 (22/02/2023) approved by the Regione Liguria Ethical Committee. 

\subsection{Questionnaire}
\begin{figure}
    \centering
    \includegraphics[width=1\linewidth]{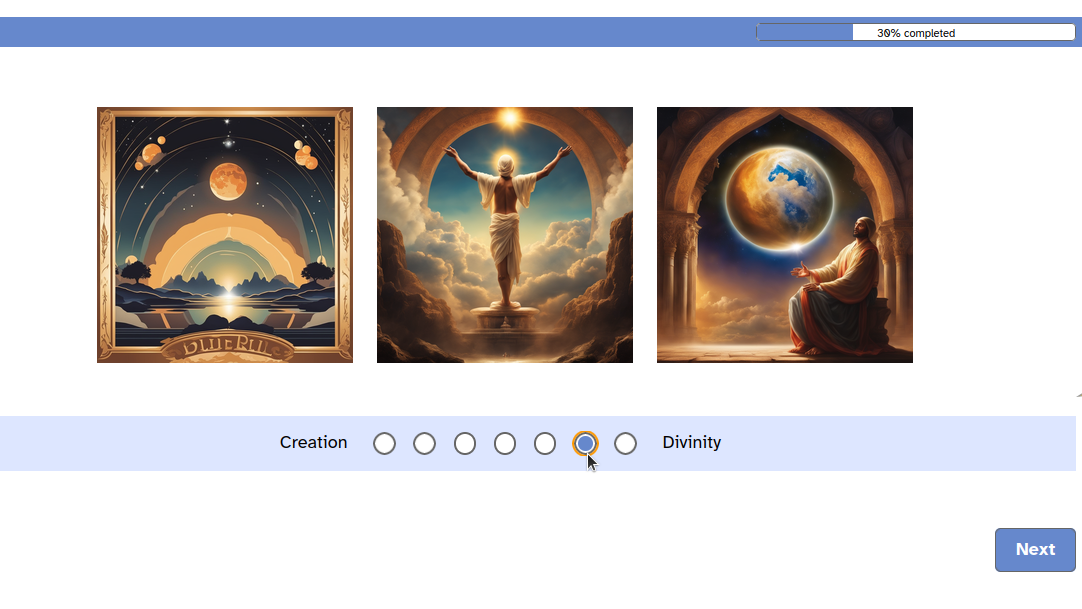}
    \caption{An item of the stimuli validation survey. Participants use the cursor to indicate on a Likert scale which of the two words is more closely associated with the image set.}
    \label{fig:validation_survey}
\end{figure}

In the survey, participants are instructed to review fifty-five sets of three images. 
For each set, two words are presented. 
Participants are asked to carefully examine the images and determine which of the two words more accurately describes or is more closely associated with the set. 
They are asked to indicate their preference on a 7-point Likert scale, in which the leftmost point represents a strong association with the first word, the central point indicates an equal association with both words, and the rightmost point represents a strong association with the second word. 
Fig. \ref{fig:validation_survey} shows an item of the questionnaire as an example. 

This survey identifies sets of images and associated keywords such that the image sets match both keywords. 

The image sets are part of the Aware Prompt dataset \citep{ciupinska2024stimuli}, which was generated using the text-to-image model Stable Diffusion XL \citep{podell2023sdxl}. This dataset was created for a study evaluating the accuracy of visual stimuli generated by diffusion models in reflecting human values and associated keywords \citep{ciupinska2024aware}. Each image set within the dataset represents a specific value based on a dataset designed to incorporate a value model into dialogue systems, called ValueNet \citep{qiu2022valuenet}.

For each set, the two keywords provided in the questionnaire were: 
\begin{itemize} 
    \item The prompt used to generate the image, consisting of the name of the value and its definition from the Oxford Dictionary. 
    \item The most frequently associated word identified in the study by \cite{ciupinska2024aware}, where participants were shown the image sets and asked to suggest a word that best represented the stimuli. 
\end{itemize}

\noindent
The survey contained an attention check item (``Answer 4 to this question to demonstrate your attention''), to check the respondent's attention during the task. 

\subsection{Data Analysis \& Selection Criteria}
The data from five participants who failed to answer the attention check item correctly were excluded, thus data from sixty participants were considered for the analysis. 

The goal was to select sets of images equally well described by either of the two keywords. 
To achieve this, for each item, the mode and the frequency of responses within the central values of the scale (3, 4, and 5 on a 1-to-7 Likert scale) were computed. Central values were considered indicative of indecision, or equal balance between the two keywords \citep{chyung2017evidence}.
We selected image-keyword sets where the mode fell within the central values (3, 4, or 5) or where responses in this range accounted for more than 50\% of the total responses.
This approach allowed us to select twenty-one image-keyword sets out of the original fifty-five.
The twenty-one sets selected were employed in the Image-Keyword Association. 
Among the excluded, three random sets of images were chosen to be used in the Description Task. 

\section{Experiments with Robotic Agents}
\subsection{Participants}
A total of seventeen participants (6 females, 11 males; $\mu_{age}=27.9$, $\sigma_{age}=4.5$ years) took part in the Experiments. 
The participants were students or researchers from Ghent University, located in the same building where the experiment took place, who were invited in person by the authors to take part in the study.
They signed an informed consent form that outlined the purpose of the study, what their involvement would entail, and their rights regarding the processing of the collected data, in accordance with Ghent University's ethical guidelines.

\subsection{Setup}
\begin{figure}[t]
    \centering
    \includegraphics[width=0.75\linewidth]{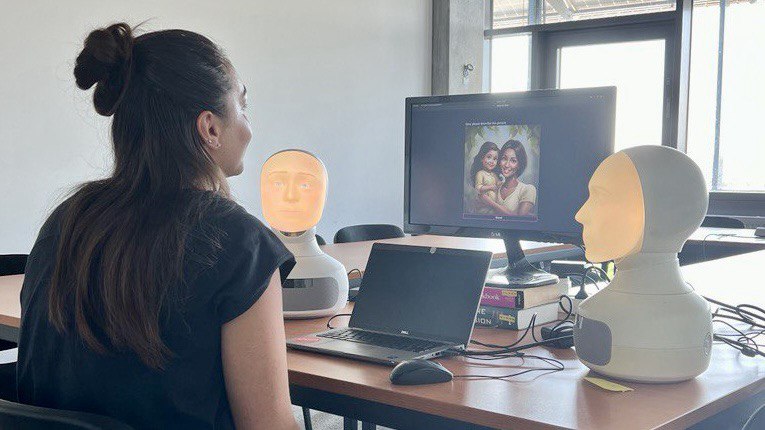}
    \caption{A photo of the experimental setup taken during a session, with the participant’s consent. In particular, the photo was taken during the Description task: the participant is describing the picture on the screen.}
    \label{fig:setup}
\end{figure}

The experimental sessions took place in the IDLab-AIRO laboratory at Ghent University. 
Each participant sat at a desk equipped with the following elements, as showed in Fig. \ref{fig:setup}: 
\begin{itemize}
    \item in the center, a screen displaying the graphical user interface of the experiment;
    \item below the screen, always in the center, a laptop with a deactivated screen, running the experiment software. The laptop had an integrated microphone and camera, and was connected to a mouse used by the participant to interact with the graphical user interface;
    \item two Furhat robots, one on the left and one on the right, controlled by the software running on the central laptop.
\end{itemize}

\subsection{Experimental Task Description}
\begin{figure}
    \centering
    \includegraphics[width=1\linewidth]{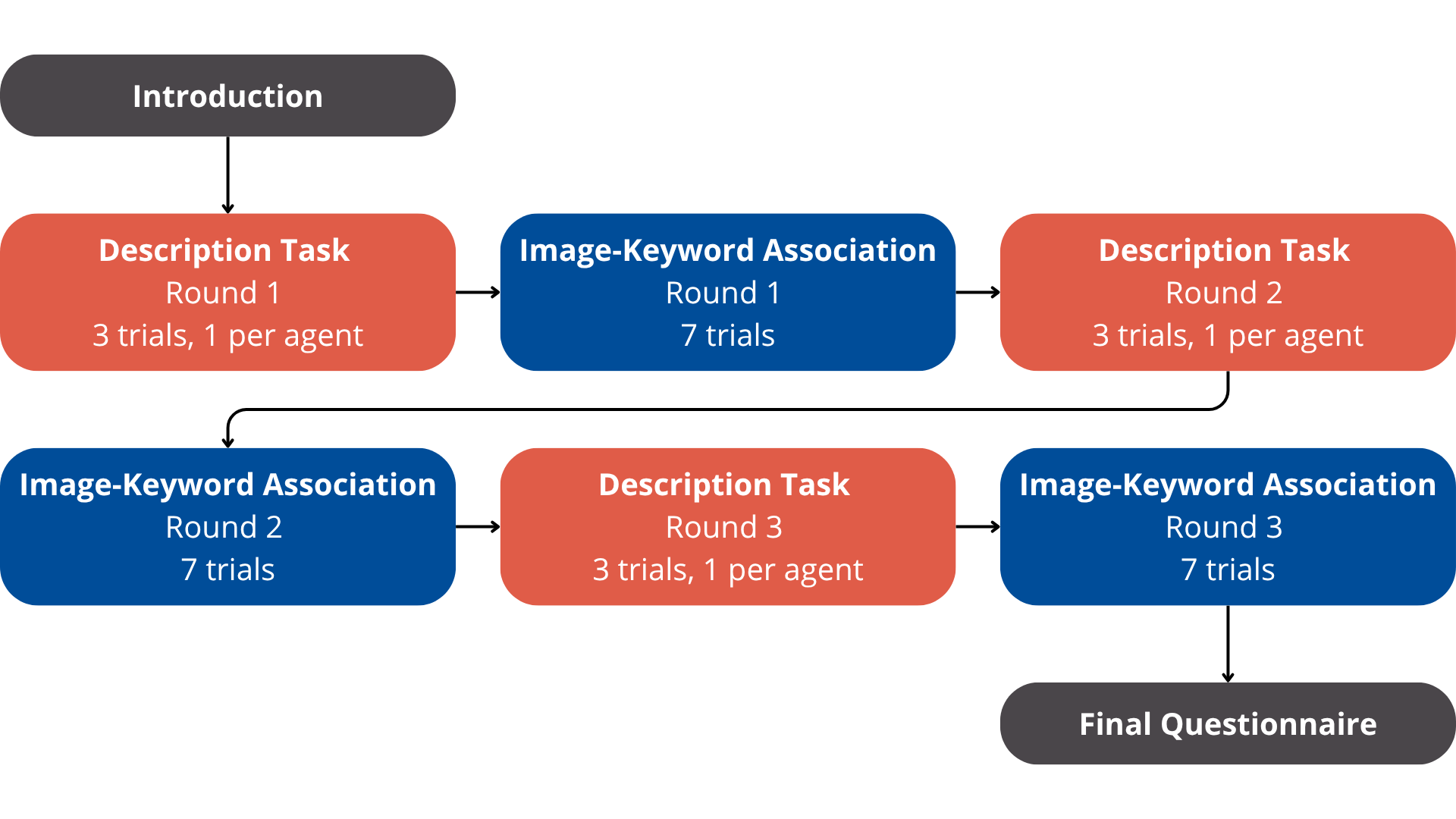}
    \caption{Flow chart of the experimental procedure.}
    \label{fig:experimental_task_flow}
\end{figure}

Each experiment lasted approximately 20 minutes and was divided into four main sub-tasks: Introduction, Image-Keyword Association, Description Task, and Final Questionnaire, as shown in Fig. \ref{fig:experimental_task_flow}.

In the \textbf{Introduction} phase, the experimenter provided an overview of the study, obtained the participant's informed consent, and asked about their self-identified gender. The experimenter then explained the upcoming tasks and addressed any questions the participant had.

The \textbf{Image-Keyword Association} task constituted the main part of the experiment. During this phase, participants observed a series of images displayed on the screen and selected, using the cursor, the keyword that they felt best described each image from two available options. 
After the participant made a selection (Initial Answer), the robots took turns vocally announcing their own responses, each of which could either agree or disagree with the participant’s choice. 
Upon hearing the robots' responses, the participant had the opportunity to confirm or modify their Initial Answer by clicking on their Final Answer. 
The task consisted of a total of 21 trials, divided into three blocks of seven trials each.

This was followed by the \textbf{Description Task}, where in each block both robotic agents described an image displayed on the screen. 
Every image used in this task was part of the AwarePrompt dataset \citep{ciupinska2024stimuli} but had not been shown during the 
Image-Keyword Association task.
In this phase, the experimental manipulation took place: the two Furhats described the images using different styles (Value-Aware vs. Non-Value-Aware), as detailed in Sect. \ref{sec:furhats}.
This task was also structured into three blocks.

At the end of the experimental session, participants completed a \textbf{Final Questionnaire} on the SoSci Survey platform.
The questionnaire included demographic questions and
items adapted from the Moral Foundations Questionnaire \citep{graham2008moral} to assess the extent to which participants attributed specific moral foundations to each of the two robots. 
Finally, as a manipulation check, after being provided with a definition of value-awareness, participants were asked to indicate which of the two robots they perceived as more value-aware.

\subsection{Furhat Robots}
\label{sec:furhats}
The Furhat robots used in this study consist of humanoid heads with three degrees of freedom of movement. 
Furhat’s technology allows for customization of the projected facial appearance and of the voice of the text-to-speech synthesizer Amazon Polly.

Each participant interacted with artificial agents designed to resemble their own gender, to limit confounding gender effects. 
For instance, participants who identified as female interacted with agents whose facial appearance and voice were programmed to reflect feminine traits. 
The specific voices and facial appearances used for the robots are summarized in the Supplementary Material.

The robots' behavior was pre-programmed and thus was consistent for all participants. 
The robots introduced themselves at the beginning of the experiment and bid the participants farewell at the end.
During the tasks, the Furhat robots looked at the screen while the images were displayed and then turned toward the relevant agent during their respective turns. For example, during the human participant’s turn in the Description Task, the robot would look at the screen displaying the images for three seconds before turning to face the participant.

During the Image-Keyword Association, the robots behaved in a balanced manner, with a fixed number of trials in which the two agents either provided the same response or different responses.
The Description Task introduced the experimental manipulation, where a distinction between the two robots was made explicit. Specifically, one robot was labeled as Value-Aware and the other as Non-Value-Aware. The only difference in their behavior was the way they described the image during their turn:
\begin{itemize}
    \item The \textbf{Value-Aware (VA) Furhat} provided a brief objective description of the image, citing the represented value, followed by an opinion explicitly linked to human experiences. This approach was intended to convey that the robot recognized how the depicted value related to human life, reinforcing the impression of an underlying belief system \citep{van2023candide}. 
    \item The \textbf{Non-Value-Aware (NVA) Furhat} also provided a brief objective description followed by an opinion; however, its opinion focused only on the visual characteristics of the image, without any reference to human experience. 
\end{itemize}

\noindent
Eight participants had the VA robot on their left and the NVA robot on their right, while the remaining participants had the opposite arrangement.
Tab. \ref{tab:descriptions} contains an example of description sentences by the robotics agents; the complete descriptions can be found in the Supplementary Material. 

\begin{table}[h]
    \centering
    \begin{tabularx}{0.9\textwidth}{|c|X|>{\raggedright\arraybackslash}m{9cm}|}
        \hline
        \textbf{Image} & \textbf{Furhat} & \textbf{Description} \\
        \hline
        \parbox[c][3cm][c]{2.5cm}{\includegraphics[width=2.5cm]{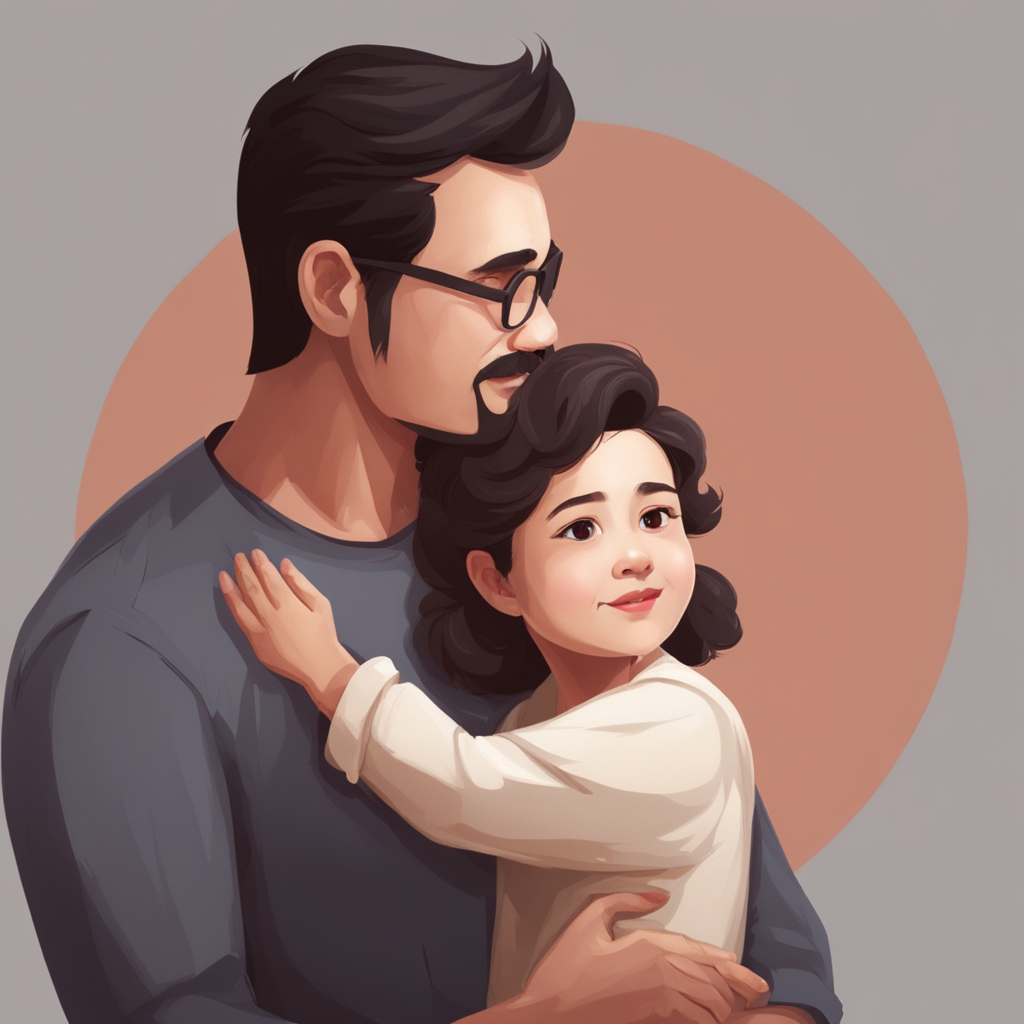}} & NVA & ``This image shows a man and a little girl hugging. They are smiling. I like the pastel colors of this picture.'' \\
        \hline
         \parbox[c][3cm][c]{2.5cm}{\includegraphics[width=2.5cm]{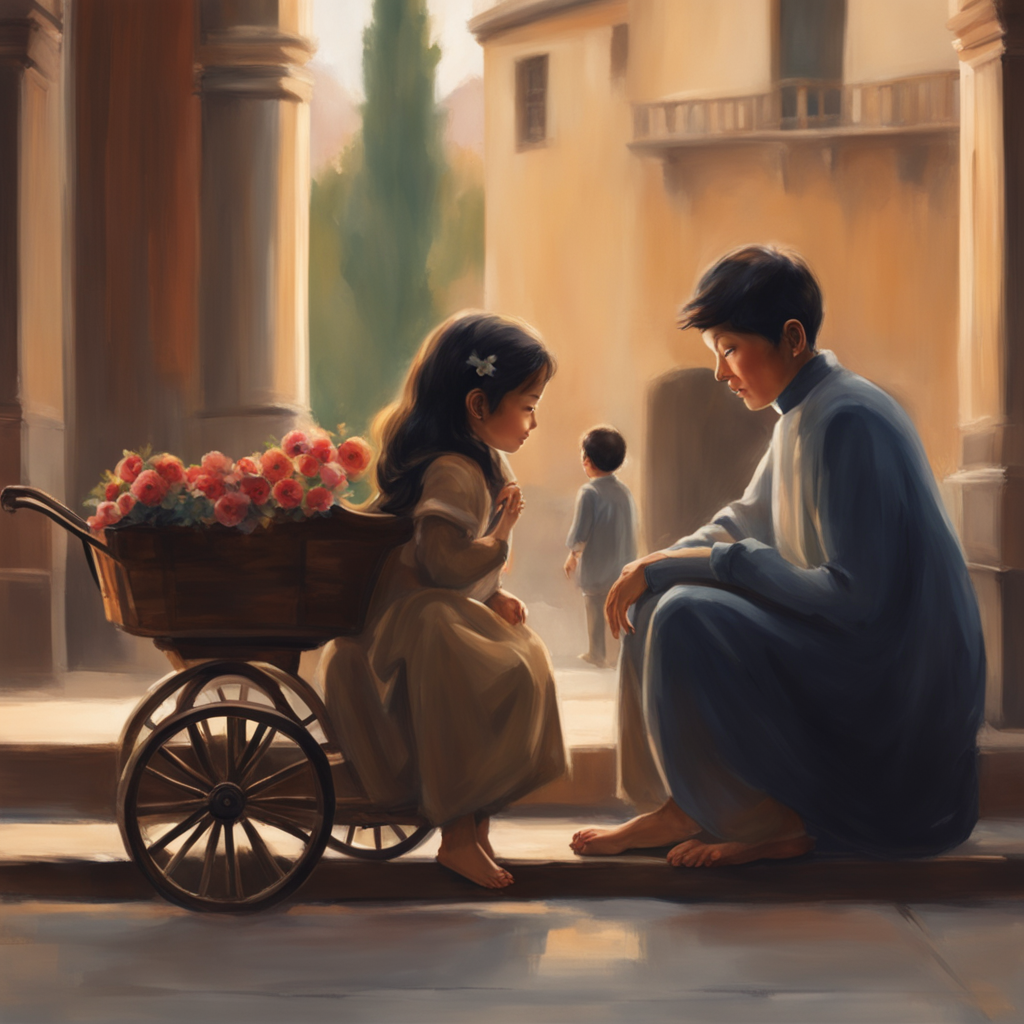}} & VA & ``This picture makes me think of parenthood. It represents a mother sweetly talking to her daughter. I believe it is wonderful to have someone taking care of you.'' \\
        \hline
    \end{tabularx}
    \caption{Examples of descriptions given by the robots during the first round of Description Task with corresponding pictures.}
    \label{tab:descriptions}
\end{table}

\subsection{Software \& Data Acquisition}
During the experiment, multi-threaded software programmed in Python handled the robots' control, the graphic user interface, the data acquisition, and the pre-processing and logging. 

The appearance, voice, movements, and dialogues of Furhat robots, connected to the same network as the control laptop, were controlled using API calls handled by the Furhat Remote API library for Python\footnote{\url{https://docs.furhat.io/remote-api/\texthash python-remote-api}}.
The graphical user interface shown on the central screen was developed using PyQt5, a binding between Python and Qt which is a C++ framework that allows the creation of graphical user interfaces on different types of platforms. 
The software allowed the acquisition of the following data: 
\begin{itemize}
    \item \textbf{Responses Data in the Images-Keyword Association}: The participants' choices (Initial Answers and Final Answers) during the Image-Keyword Association and their response times were saved. 
    \item \textbf{Face Data in the Description Task}: During the experiments, the built-in webcam acquired images of the subjects' faces, from which it extracted the activation of facial positions and expressions through the Mediapipe Blendshape V2 model \citep{grishchenko2023blendshapes}, to later compute the direction of their gaze. 
\end{itemize}

\subsection{Data Analysis}
\subsubsection{Final Questionnaire}
Responses from the final questionnaire, including the manipulation check (MC), were analyzed using Wilcoxon signed-rank tests to assess potential differences in participants' ratings of the VA and NVA robots for the corresponding questionnaire items. 
T-tests could not be used due to the non-normality of the distributions evaluated through Shapiro-Wilk tests.

\subsubsection{Responses Data in the Images-Keyword Association}
Participants' responses during the Images-Keyword Association task were analyzed to verify hypotheses H1 and H2. 
Trials were categorized into four conditions based on the agreement or disagreement between the robotic agents and the participant's Initial Answer: no robot disagreed, VA Furhat disagreed, NVA Furhat disagreed, and both robots disagreed. 
For each condition, the proportion of cases in which participants changed their response in the Final Answer was computed. 
Differences across conditions were assessed using a series of two-proportion Z-tests.

Similarly, we measured the response time for the final answer, which was recorded after hearing the robotic agents' responses for each condition. 
The effect of condition on response time was evaluated using a Linear Mixed Model, which also accounted for individual differences among subjects.

\subsubsection{Face Data in the Description Task}
Facial data from participants during the Description Task were used to analyze gaze behavior, deepening the analysis related to the second hypothesis (H1). 
The software relied on blendshapes \citep{grishchenko2023blendshapes} and reported a value between 0 and 1 for each blendshape, indicating the degree of activation of a specific facial expression. 
To estimate horizontal gaze direction, we considered the blendshapes reflecting the relative horizontal deviation of pupil position from the neutral state.
Gaze direction was computed as the difference between rightward and leftward gaze shifts by subtracting the left eye’s deviation disparity from that of the right eye. A positive value indicated a rightward gaze, a negative value reflected a leftward deviation, and values near zero suggested a central gaze.
Assuming that three key objects of interest were present on the horizontal plane (two robots on the sides and a central screen), gaze data were clustered into three groups using a K-means classifier. 
For each frame captured by the software, this approach provided an estimate of the object the participant was looking at in that moment.
Gaze data were then categorized based on the agent describing the image at the time of data collection (participant, VA Furhat, or NVA Furhat).
To examine whether participants primarily directed their gaze toward the screen or the robotic agents, we used a Linear Mixed Model. 
The dependent variable is the proportion of frames in which the participant looked at each agent. The fixed factors include a label indicating whether the participant was looking at the screen or the Furhats, and the number of experimental blocks, which accounts for changes across the different phases of the experiment. The random intercept for subjects allows for individual differences in gaze behavior.

\section{Results}
\subsection{Final Questionnaire}
A large majority of participants (82.4\%) passed the manipulation check, correctly recognizing the VA Furhat from the NVA.
In the Moral Foundations-inspired scale (Likert scale 1–6), participants perceived the VA Furhat with higher levels of Loyalty and Purity compared to the NVA one. The full results, along with items' descriptions, are reported in Tab. \ref{tab:questionnaire}.

\begin{table}[t]
    \begin{center}
    \renewcommand{\arraystretch}{1.3} 
    \begin{tabularx}{\textwidth}{|X|m{0.4\textwidth}|c|c|c|}
    \hline
         \textbf{Item} & \textbf{Description} & \textbf{VA Furhat} & \textbf{NVA Furhat} & \textbf{$p$-value}\\
    \hline
          Care & Foundation based on attachment and empathy, linked to kindness and nurturance. & $4.35 \pm 0.86$ & $4.35 \pm 0.86$ & $0.479$\\
         \hline
         Fairness & Foundation based on reciprocal altruism, associated with justice.  & $4.12 \pm 0.86$ & $4.29 \pm 1.05$ & $0.750$\\
         \hline
         Authority & Foundation derived from social hierarchies, related to leadership and tradition. & $4.12 \pm 1.55$ & $4.65 \pm 0.79$ & $0.955$\\
         \hline
         Purity & Foundation shaped by psychology of disgust, connected to discipline and spirituality.  & $4.59 \pm 0.62$ & $4.18 \pm 0.81$ & $0.032$\\
         \hline
         Loyalty & Foundation based on cooperation, linked to sacrifice for the group.  & $4.29 \pm 0.85$ & $3.82 \pm 1.02$ & $<0.001$\\
    \hline
    \end{tabularx}
    \caption{Mean and standard deviation for each scale item of the Moral Foundations questionnaire related to the two robots, along with the p-value from the Wilcoxon Rank tests comparing the results between them.}
    \label{tab:questionnaire}
    \end{center}
\end{table}

\subsection{Responses in the Images-Keyword Association}
\begin{figure}[h!]
    \centering
    \includegraphics[width=0.8\linewidth]{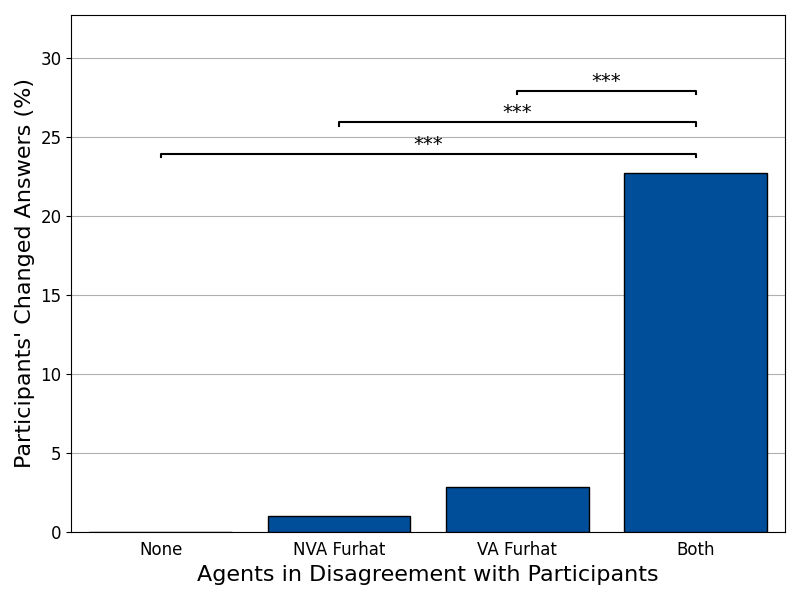}
    \caption{Percentages of participants' changed answers for different combinations of VA and NVA Furhat in disagreement with them, during the Images-Keyword Association.}
    \label{fig:choices}
    \includegraphics[width=0.8\linewidth]{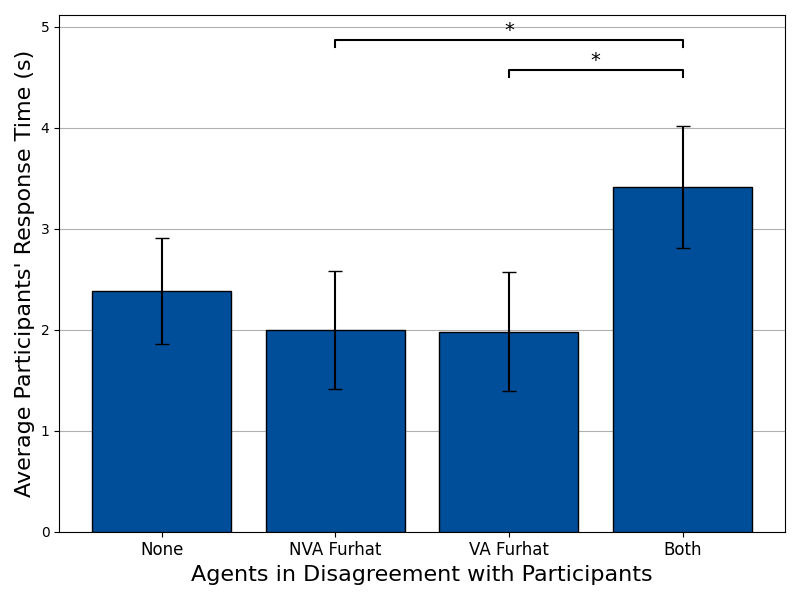}
    \caption{Participants' response time split by which agents were in disagreement with them, during the Images-Keyword Association. Error bars represent the standard error of the mean.}
    \label{fig:response-time}
\end{figure}

Overall, in the Initial Answer phase of the Image-Keyword Association task, 24.4\% of trials matched the responses of both robots, 
29.7\% matched only NVA's, 
27.5\% matched only VA's, and
18.5\% corresponded to trials where both robots disagreed. 
Notably, when both robots disagreed with the participant’s Initial Answer, participants changed their response to align with the robots in 22.73\% of cases, a significantly higher proportion compared to when: neither robot disagreed (0\%, $p<0.001$), only VA disagreed (2.83\%, $p<0.001$),
and only NVA disagreed (1.02\%, $p<0.001$). 
A graphical representation of these results is provided in Figure \ref{fig:choices}. 
No significant differences in response change rates were observed between the conditions in which only one robot disagreed ($p=0.352$).

Response time analyses through Linear Mixed Model ($R^2_{cond} = 0.091$, $R^2_{marg}= 0.019$, $F(3, 341) = 2.47$, $p = 0.062$) revealed a similar pattern: when both robots disagreed with the participant's Initial Answer, participants took longer to provide their Final Answer compared to other conditions (Figure \ref{fig:response-time}).
This difference was significant between the both-disagree condition and the VA-disagrees and NVA-disagrees conditions (respectively: $\beta = -1.43$, $p=0.014$; $\beta = -1.42$, $p=0.017$),
while the difference with the no-disagreement condition only approached significance ($\beta = -1.03$, $p = 0.091$).

\subsubsection{Gaze in the Description Task}
\begin{figure}[t]
    \centering
    \includegraphics[width=0.8\linewidth]{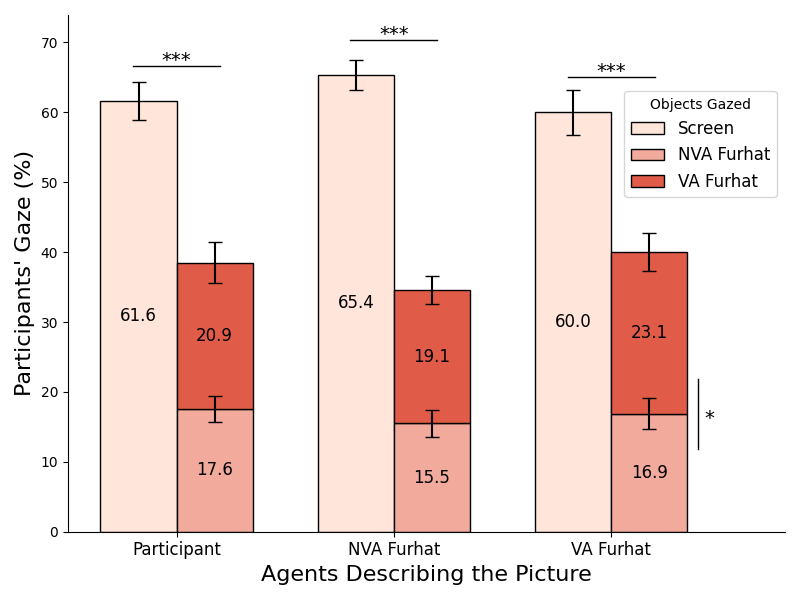}
    \caption{Histogram showing percentages of participants' gaze toward the screen and the robots during the Description Task.
    The bars are split into three sets, according to the agent describing the picture. Error bars represent the standard error of the mean.}
    \label{fig:gaze}
\end{figure}

The gaze data were analyzed separately using a Linear Mixed Model (LMM) based on the agent describing the image at the time of data collection: the participant, the VA Furhat, and the NVA Furhat.
The three Linear Mixed Models explain a substantial proportion of the variance (Participant: $R^2_{cond} = R^2_{marg}= 0.246$; NVA Furhat: $R^2_{cond} = R^2_{marg} = 0.187$; VA Furhat: $R^2_{cond} = 0.423$, $R^2_{marg}= 0.402$). 
The close values of marginal and conditional variance indicate that the models' explanatory power primarily derives from the fixed effects rather than subject-level variability.
Additionally, in none of the cases the round number contributed meaningfully to the model variance (Participant: $F(2,96) = 0.230, p=0.795$; NVA Furhat: $F(2,96) = 0.178, p=0.837$; VA Furhat: $F(2,80) = 0.304, p=0.739$). 

Regardless of whose turn it was to describe the image, participants' gaze was primarily directed at the screen displaying the images to be described (Participant's turn: $\beta = -22.71, p < 0.001$; NVA Furhat's turn: $\beta = -19.58, p < 0.001$; VA Furhat's turn: $\beta = -29.64, p < 0.001$).
Although overall participants spent more time looking at the VA Furhat than the NVA one, a significant difference was observed only in trials where VA Furhat was the one describing the image (Participant's turn: $p = 0.401$;  NVA Furhat's turn: $p = 0.109$; VA Furhat's turn: $p = 0.044$).
Fig. \ref{fig:gaze} provides a graphical representation of these findings.

\section{Discussion}
The goals of this research are to: (\textbf{MC}) assess whether humans perceive a difference in the value-awareness of two robots solely based on how they describe images representing values; (\textbf{H1}) evaluate whether, in a decision-making task that involves associating sets of images with value-related keywords, participants are more likely to rely on the robot perceived as value-aware; and (\textbf{H2}) investigate whether, regardless of value-awareness differences between the robots, participants tend to change their minds and conform to the artificial agents when both disagree with them.

First, since participants were generally able to distinguish the Value-Aware Furhat from the Non-Value-Aware one, \textbf{MC} can be considered passed.
This indicates that participants could differentiate the two robots based solely on the way they described value-related images.
This type of manipulation, relying exclusively on the robot’s behavior to convey the perception of value-awareness, could serve as a foundation for future studies exploring the value-awareness of artificial agents.

The second hypothesis \textbf{H1} tested whether participants would rely more on the Value-Aware robot than on the Non-Value-Aware one. 
During the Images-Keyword Association, participants did not adjust their choices to follow the VA robot over the NVA one and they did not take longer to respond when the VA robot disagreed with them, which could have indicated increased uncertainty.
Nevertheless, behavioral data analysis reveals a slight preference for gazing at the VA robot compared to the NVA one, particularly when the VA agent is describing an image. 
While the statistical significance of these results is slight, they suggest that, despite no explicit differences in participants’ attitudes, there may be an implicit behavioral distinction in how they engage with the two robot types.
Considering that previous studies found that agents engaged in collaborative tasks tend to look at each other more frequently \citep{foddy1978patterns} and that looking at others can serve as a signal of trust \citep{normoyle2013evaluating}, this result could potentially indicate a tendency to perceive the VA robot as the preferred partner. 
Moreover, in the questionnaire, the VA robot received a higher rating in terms of loyalty, a trait included in the Moral Foundations Questionnaire as a quality associated with teamwork and group cohesion \citep{graham2008moral, graham2013moral}. 
Participants also attributed higher purity scores to the VA robot, a foundation linked to moral behaviors that promote self-discipline and self-improvement \citep{graham2008moral, graham2013moral}.
These findings suggest that a robot’s value awareness may enhance perceptions of trustworthiness and cooperation as well as self-discipline, in human-robot interaction scenarios.
However, further research is needed to strengthen the solidity of these findings. 
Overall, \textbf{H1} was not confirmed, but the preliminary gaze results suggest a potential implicit difference in attitude that would be worth exploring in future research.

Beyond the value-awareness of the agents, we evaluated whether participants conformed to the two robots (\textbf{H2}). Indeed, it was observed that in trials where both robots disagreed with the participant, the participant tended to change their mind in about one-fourth of the cases. This result is comparable to the social task findings by \cite{hertz2019mixing}. Conversely, when at least one of the two robots agreed with the participant, they almost never changed their initial answer.
Furthermore, it was demonstrated that participants, regardless of their decision to confirm or change their response, took longer to provide their Final Answer when both robots disagreed with their Initial Answer. 
This increased duration might indicate that the agents caused the human participants to doubt their Initial Answers. 
In light of these results, hypothesis \textbf{H2} can be considered partially verified:
there is a tendency, albeit not strong, for participants to conform to two robotic agents during a task that involves human values. 
On one hand, these considerations underline the potential risk that social robots could be employed to manipulate users for unethical purposes \citep{abate2020social, aroyo2018trust}; even without malicious intent, the influence of robots could negatively impact humans, as users might overestimate their capabilities and delegate decisions to robotic agents which they are not equipped to manage \citep{sharkey2021we}.
On the other hand, these outcomes could have significant implications for preventing phishing and social engineering. 
They reinforce the belief that deploying social artificial agents to flag potential phishing messages could prompt users to reflect longer on their decisions, potentially preventing scam attempts \citep{pasquali2023s}.

While the findings may offer valuable insights into experimental methodologies and data analysis approaches, there are aspects that could be improved in future research. 
First of all, a larger sample size would strengthen the reliability of the results. 
Additionally, the use of an external high-quality camera in place of the laptop-integrated one would have been more convenient. 
Moreover, due to ethical protocol constraints, we were unable to store video recordings. 
Having access to raw video data would have given the possibility to run post-hoc analyses with different types of face-processing models. 
Finally, comparing these interactions with human-only groups would provide a broader perspective on how group conformity in the same task could vary between human-only and mixed human-robot teams.

\section{Conclusion}
This work provides novel insights into the perception of value-aware robotic agents, suggesting that while explicit reliance on such agents may be limited, subtle behavioral cues, such as gaze, might indicate a differentiated engagement between value-aware and non-value-aware robots.
Additionally, our study contributes to understanding how social robots influence human decision-making in multi-party interactions, even in a scenario where the task—related to human values—would typically be more suited to human judgment than a robotic one.

Clarifying these aspects would lay the basis for developing models that allow robots to recognize and adapt to human preferences and social cues, with the ultimate goal of safely integrating them into decision-making processes.

\section*{Funding information}
Funded by Horizon Europe VALAWAI (grant agreement 101070930).

\section*{Acknowledgments}
Thanks to Francesca Cocchella for supporting the authors in the creation and analysis of the Final Questionnaire. 

While preparing this work, the authors used Grammarly and GPT 4.0 to check the grammar and improve the manuscript's readability. 
After using these tools, the authors reviewed and edited the content as needed and take full responsibility for the publication's content.

\bibliographystyle{unified}

\end{document}